\documentclass[a4paper]{Springer_style/svproc}

\usepackage{url}
\usepackage{graphics}
\usepackage{epsfig}
\usepackage{graphicx}
\usepackage{amsmath}
\usepackage{amssymb}
\usepackage{algorithm}
\usepackage{algorithmic}
\usepackage{booktabs}
\usepackage{multirow}
\usepackage{subcaption}
\usepackage{xcolor}
\usepackage{float}

\newlength{\imgheight}

\begin{document}
\mainmatter

\title{Gripper-Aware Automatic Dense Packing of Irregular Objects}
\titlerunning{Dense Packing}

\author{Tianhao Qin\and Connor McCann \and Berk Calli \and Jing Xiao}
\authorrunning{Tianhao Qin et al.}
\tocauthor{Tianhao Qin, Connor McCann, Berk Calli, and Jing Xiao}
\institute{Robotics Engineering Department\\
Worcester Polytechnic Institute, Worcester, MA 01609, USA\\
\email{tqin@wpi.edu}}

\maketitle

\begin{abstract}
Automatic dense packing is widely desired in warehouse operations but remains a fundamental challenge in robotic manipulation. Existing work on irregular-object packing largely targets simulation with idealized contact, treating the object as an isolated rigid body. The gripper often enters as a discrete, post-hoc feasibility check, if considered at all, and the perception and contact drift accumulated during execution are not addressed. We present a closed-loop pipeline that integrates perception, gripper-aware placement optimization, and force-guided execution on a real manipulator. The optimizer represents the object together with the gripper as a single composite body of hierarchical sphere trees. It searches over five degrees of freedom on a GPU within a CMA-ES framework, with the vertical coordinate grounded analytically against the current heightmap. During execution, a force-monitored vertical descent stops on first contact. A post-release consolidation push then closes the residual lateral clearance that gripper-aware planning leaves behind. The container is re-perceived between placements so that drift does not accumulate. We validate the system on a Franka Emika Panda robot packing a 3D-printed set of flat, curved, and concave objects, and a YCB object subset. An ablation study isolates the contribution of gripper-aware optimization, the consolidation push, and mesh-derived geometry to end-to-end success, achieved density, and computational cost. We further benchmark against the heightmap-minimization method as a baseline representative of prior irregular-object packing work.
\end{abstract}

\section{Introduction}
\label{sec:intro}

Dense packing of objects into a container is a common task in warehouse logistics and order fulfillment, where performance depends on how efficiently each container is filled. Classical formulations, dating back to bin-packing and pallet-loading, assume the items are regular boxes with known dimensions~\cite{martello2000threedimensional}. However, such formulations no longer match today's operations that increasingly require densely packing irregular items. Recent work has accordingly turned to packing irregular objects, but the gap still remains when those approaches are executed on a real robot. In this paper, we present a full robotic system capable of dense packing of objects with arbitrary shape. 

\subsection{Related Work}
\label{sec:related}

\subsubsection{Irregular-Object Packing.}

Automatic packing of irregular objects has progressed much in recent years. Wang and Hauser introduced a heightmap minimization formulation over arbitrary meshes for stable bin packing with a robot manipulator~\cite{wang2022densepacking}. However, it is executed open-loop, with the object treated as an isolated rigid body. Learning-based methods have also been applied to bin-packing. In the cuboid setting, PackerBot~\cite{packerbot2021} uses heuristic-assisted deep reinforcement learning, and Packing Configuration Tree~\cite{zhao2022pct} learns over a tree of candidate placements. Both achieve strong densities on cuboid items but do not extend to arbitrary mesh geometry. For irregular shapes, Huang et al.~\cite{huang2023hrl} use hierarchical reinforcement learning to jointly plan packing sequence and placement, IR-BPP~\cite{zhao2023irbpp} combines geometric candidate generation with a learned placement-selection policy trained in physics simulation, and SDF-Pack~\cite{pan2023sdfpack} formulates placement as signed-distance-field minimization. Most recently, RoboPacker~\cite{wu2025robopacker} integrates shape estimation, hierarchical reinforcement learning for packing sequence and placement, and manipulation strategies into an autonomous system. These methods report strong packing performance in simulation, and several also demonstrate physical robotic packing. However, the object is planned as an isolated rigid body, contact models are idealized, and gripper geometry never enters as part of a continuous placement objective. We hypothesize that in dense configurations the gripper geometry is equally if not more likely to collide with the environment than the object itself, so a placement that fits the object alone may still be unreachable. 

\subsubsection{Execution-Aware and Contact-Rich Placement.}

A separate line of work addresses the gap between simulation and execution. Even with a heightmap planner, Wang and Hauser report that open-loop physical execution succeeds on only 83\% of five-item orders without resensing and replanning after a placement shifts~\cite{wang2022densepacking}, indicating that strong simulation-time density alone does not translate to reliable real-robot performance. This gap has several sources that prior work has tackled in isolation. Ignoring gripper geometry in the planner produces approach-time collisions, perception drift between planned and achieved poses accumulates across an open-loop sequence, and small lateral errors that would be harmless in free space cause the object to wedge when seated into a tight gap. Conservative planning that enlarges clearances around each placement trades achievable density against the risk of an infeasible execution~\cite{wang2022densepacking}, while hand-designed primitives such as toppling and push-to-place repair the configuration after the nominal placement~\cite{shome2019packing}. Recent benchmarking~\cite{robobpp2025} quantifies the gap between geometric compactness and physics-feasible execution in simulation. Most closely related, Cao and Xiao~\cite{cao2026tase} present a complete pipeline for online dense packing of novel objects that reasons about sensing and motion uncertainty, but the objects are restricted to prism-like shapes that are irregular only in planform, and a vacuum gripper is used mostly to simplify the effect of grasping in packing optimization. What is still missing is a closed perception-to-execution loop that couples gripper-aware placement optimization with contact-reactive descent so that drift accumulated across a sequence of placements is absorbed at the time scale it appears. A related family of contact-rich assembly methods~\cite{cao2024ral} couples a sphere-tree representation~\cite{xiao2013haptics} with active force/torque-guided recovery. Their part-into-part setting is different from packing, but the sphere-tree representation is a direct precedent for ours.

\subsection{Approach and Contributions}

This paper addresses dense packing of irregular objects with a soft parallel jaw gripper, under the perception and contact uncertainties that are often ignored in simulation-optimal pipelines. We assume a pre-registered library of object meshes with associated grasps, in line with industrial settings where a unit catalog is available. 

The contributions of this paper are as follows:
\begin{itemize}
    \item A \emph{gripper-aware placement optimizer} that reasons about the object and the gripper as a single composite body and searches over 5 DoF on a GPU.
  
    \item A \emph{closed-loop execution layer} that absorbs perception and contact drift through force-monitored descent and a post-release consolidation push, with the container re-perceived between placements.
  
    \item \emph{System validation} on real hardware over a set of 3D-printed and YCB objects with various shapes, with ablations isolating the contribution of each component and comparative study against a prior baseline.
  
\end{itemize}

\vspace{-1em}
\begin{figure*}[h]
\centering
\includegraphics[width=\textwidth]{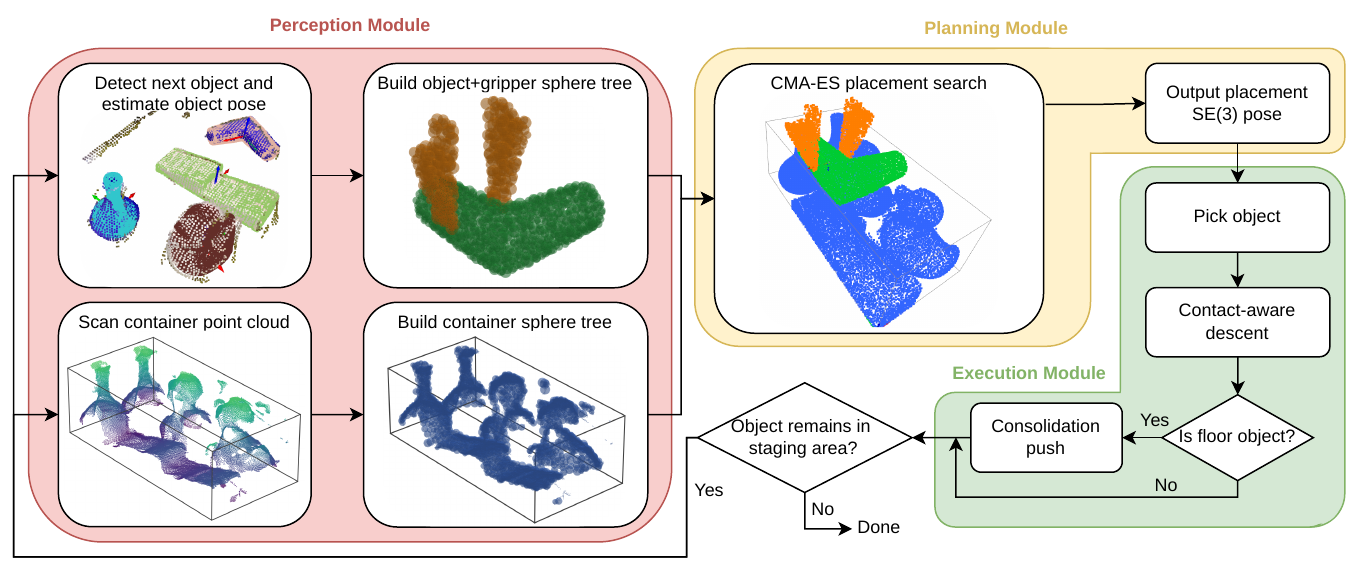}

\caption{Dense packing pipeline. Perception detects the next target object and scans the container, Planning composes the object with its grasp and searches over placements with CMA-ES, and Execution picks, descends with force/torque monitoring, and consolidates eligible placements before looping back.}
\label{fig:system_overview}
\end{figure*}

\section{Method}
\label{sec:method}

Figure~\ref{fig:system_overview} shows the overall pipeline. At iteration $k$, the system perceives the workspace to obtain the current container configuration $\mathcal{C}_k$ and the next target object, runs the gripper-aware placement optimizer over $\mathcal{C}_k$ and the target object (Sec.~\ref{sec:opt}), and executes the resulting placement with force-monitored descent and post-release consolidation (Sec.~\ref{sec:exec}) before looping back for $\mathcal{C}_{k+1}$.

\subsection{Gripper-Aware Placement Optimization}
\label{sec:opt}

The optimizer treats the object and the gripper holding it as a single composite body, and searches over five degrees of freedom under a multi-term objective that is evaluated in batches on a GPU.

\subsubsection{Composite Body and 5-DOF Grounded Formulation.}

Let $P_O \in \mathbb{R}^{N_O \times 3}$ be the target object point cloud in the object's own frame, and $P_G \in \mathbb{R}^{N_G \times 3}$ be the gripper point cloud, rigidly attachedthrough the pre-registered grasp. The union $P_O \cup P_G$ forms the composite body. A candidate placement pose is a rigid-body transform in the workspace frame whose origin is the container corner and whose axes are parallel to the robot base frame, with position $(x,y,z)$ and an axis-angle rotation $\boldsymbol{\theta} \in \mathbb{R}^3$. The search vector is defined as
\begin{equation}
\mathbf{x} \;=\; (x,\, y,\, \theta_1,\, \theta_2,\, \theta_3) \;\in\; \mathbb{R}^5
\label{eq:searchvec}
\end{equation}
excluding $z$. For each \textbf{x} sampled by the search, $z$ is set analytically as the highest support point beneath the object, by indexing a heightmap $h$ of the current $\mathcal{C}_k$ at 5\,mm cell resolution:
\begin{equation}
z^*(x, y, \boldsymbol{\theta}) \;=\; \max_{(i,j) \in \mathcal{N}_O(x,y)} h(i,j) \;-\; \min_{p \in P_O} \bigl[R(\boldsymbol{\theta}) \, p\bigr]_z,
\label{eq:zground}
\end{equation}
where $\mathcal{N}_O(x,y)$ is the set of heightmap cells above the rotated object's projected footprint. The first term is the height of the tallest surface already present beneath the object at position $(x, y)$, whether the container floor or a previously placed object. The second term is the object's lowest point once it has been rotated by $R(\boldsymbol{\theta})$, measured about its own origin. Subtracting the latter from the former drops the rotated object straight down until it rests on that support, so the placement neither floats above the configuration nor interpenetrates it. Operation~\eqref{eq:zground} is a discrete max over a finite cell set, not a differentiable closed form. CMA-ES does not require gradients, so the discontinuity is benign, and floating placements that are not physically possible are ruled out at no extra search cost.

\subsubsection{Sphere-Tree Representation.}

For each candidate, the optimizer queries distances and penetrations between the composite body and the current container configuration $\mathcal{C}_k$. To keep these queries fast, we represent each cloud as a hierarchical sphere tree~\cite{quinlan1994spheretree}, built by recursive PCA-based binary splitting. At each node we fit a Ritter bounding sphere~\cite{ritter1990efficient}, and we stop recursing once a leaf holds ten points. We write $\mathcal{S}(\cdot)$ for the sphere-tree representation of a cloud, so $\mathcal{S}_k=\mathcal{S}(\mathcal{C}_k)$ is the container sphere tree at iteration $k$ and $\mathcal{S}_O, \mathcal{S}_G$ are built once per object and gripper. The leaf centers and radii of the moving composite body are then transformed by the candidate $(x,y,z^*, \boldsymbol{\theta})$ and tested against the leaves of $\mathcal{S}_k$ in parallel as tensor operations.

\subsubsection{Multi-Term Objective.}
\label{sec:objective}

We score a candidate $\mathbf{x}$ by a weighted sum of eight terms,
\begin{align}
f(\mathbf{x}) = \;& w_\mathrm{vol}\hat{f}_\mathrm{vol} + w_\mathrm{cen}\hat{f}_\mathrm{cen} + w_\mathrm{pck}\hat{f}_\mathrm{pck} + w_\mathrm{hgt}\hat{f}_\mathrm{hgt} + w_\mathrm{fro}\hat{f}_\mathrm{fro} \nonumber\\
&+ w_\mathrm{col}\hat{f}_\mathrm{col} + w_\mathrm{bnd}\hat{f}_\mathrm{bnd} + w_\mathrm{ori}\hat{f}_\mathrm{ori},
\label{eq:objective}
\end{align}
where each $\hat{f}_i$ is a term-specific scalar score computed from the geometries of $\mathcal{S}_k$ and the composite body. The first five are \emph{packing-quality} terms, evaluated on the object sphere tree $\mathcal{S}_O$ alone because they describe the configuration that remains after the gripper retracts. The latter three represent \emph{feasibility} terms, evaluated on the composite sphere tree $\mathcal{S}_O \cup \mathcal{S}_G$ or on $\mathcal{S}_G$ alone, because the gripper should be taken into account while it holds the object. In what follows, $C(\mathcal{S})$ denotes the set of leaf sphere centers of a sphere tree $\mathcal{S}$, and a bar denotes their centroid.

For the packing-quality terms, \emph{Volume} $\hat{f}_\mathrm{vol}$ is the axis-aligned bounding-box volume of $C(\mathcal{S}_k) \cup C(\mathcal{S}_O)$, rewarding placements that compress into a small enclosing box. \emph{Centroid distance} $\hat{f}_\mathrm{cen} = \lVert \overline{C(\mathcal{S}_k)} - \overline{C(\mathcal{S}_O)} \rVert$ pulls the new object toward the existing container configuration. \emph{Packing gap} $\hat{f}_\mathrm{pck}$ averages the distance from each container leaf center $c \in C(\mathcal{S}_k)$ to its nearest center in $C(\mathcal{S}_O)$. It penalizes empty space against the current pile. \emph{Height} $\hat{f}_\mathrm{hgt}$ penalizes both the maximum $z$ over $C(\mathcal{S}_k) \cup C(\mathcal{S}_O)$ and the $z$ of $\overline{C(\mathcal{S}_O)}$. \emph{Frontier} $\hat{f}_\mathrm{fro}$ pulls $\overline{C(\mathcal{S}_O)}$ toward the workspace origin only while floor occupancy is below 70\%, preventing it from fighting the height and packing terms once the base layer is in place.

For the feasibility terms, \emph{Collision} $\hat{f}_\mathrm{col}$ sums pairwise sphere penetrations between $\mathcal{S}_O \cup \mathcal{S}_G$ and $\mathcal{S}_k$. It is evaluated on the full composite body so the gripper itself cannot pass through placed objects. \emph{Bounds} $\hat{f}_\mathrm{bnd}$ sums per-axis container overflows with a split rule, where object spheres must stay inside at all heights, while gripper spheres only have to stay inside when below the container rim. \emph{Orientation} $\hat{f}_\mathrm{ori}$ combines a hard-cone penalty when the gripper $z$-axis tilts more than $\alpha_\text{max}=45^{\circ}$ with a soft term linearly penalizing any tilt.

\begin{table}[t]
\caption{Optimization configuration: objective function weights with point-set assignments, and CMA-ES parameters.}
\label{tab:opt_config}
\centering
\small
\setlength{\tabcolsep}{3.5pt}
\begin{tabular}{lcccccccc}
\toprule
Term      & Vol. & Cent. & Pack. & Height & Front.$^\dagger$ & Coll. & Bounds & Orient.  \\
Symbol    & $\hat{f}_\mathrm{vol}$ & $\hat{f}_\mathrm{cen}$ & $\hat{f}_\mathrm{pck}$ & $\hat{f}_\mathrm{hgt}$ & $\hat{f}_\mathrm{fro}$ & $\hat{f}_\mathrm{col}$ & $\hat{f}_\mathrm{bnd}$ & $\hat{f}_\mathrm{ori}$  \\
\midrule
Weight    & 2.0 & 2.0 & 2.0 & 10\,/\,5 & 10 & 50 & 50 & 50\,/\,4  \\
Point set & O & O & O & O & O & O+G & O+G & G  \\
\bottomrule
\end{tabular}\\[2pt]
\begin{minipage}{0.95\columnwidth}
\footnotesize
O = Object points, G = Gripper points, O+G = both. Height weight $10/5$ denotes (max-$z$ weight)\,/\,(centroid-$z$ weight); Orient.\ weight $50/4$ denotes (hard-cone weight)\,/\,(soft-tilt weight). $^\dagger$Frontier active only while floor occupancy $<70\%$. \textbf{CMA-ES:} population 128, max generations 200, initial $\sigma=1.0$, patience 50, independent restarts 3.
\end{minipage}
\end{table}

Weights and point-set assignments are summarized in Table~\ref{tab:opt_config}. The relative magnitudes encode a strict priority. Feasibility terms carry weights an order of magnitude larger than packing-quality terms, so violating a hard constraint costs far more than leaving an empty corner. Within the packing-quality group, the height term dominates so that low and flattening placements are preferred over compact placements that grow upward. The soft orientation weight is intentionally small to steer the search toward upright grasps while not overriding packing quality when a tilted grasp is genuinely better. The weights and the 70\% gate were fixed once by an offline grid search, not tuned per object or per trial. 

\subsubsection{GPU-Batched CMA-ES.}

We optimize $f$ over $\mathbf{x} \in \mathbb{R}^5$ with CMA-ES~\cite{hansen2001cmaes}, whose gradient-free, population-based search suits the non-convex, mildly multi-modal landscape of $f$ avoids the non-differentiability of Eq.~\eqref{eq:zground}. At each generation the entire population is evaluated in a single batched GPU call. To avoid settling into a suboptimal local minima, the default configuration runs three independent CMA-ES instances with distinct seeds and keeps the best result. Optimization parameters are summarized in Table~\ref{tab:opt_config}.

\subsection{Closed-Loop, Uncertainty-Aware Execution}
\label{sec:exec}

The placement returned by the optimizer is a nominal target, but not a guaranteed achievable pose. Calibration error, slip during the grasp, and contact dynamics during descent each move the achieved pose away from the planned pose. The execution layer absorbs these errors at the time scale they appear on, instead of budgeting for them as inflated clearances in the optimizer.

\subsubsection{Force-Monitored Continuous Descent.}
Guarded motion under force/torque sensing is a long-standing technique in force-controlled assembly~\cite{mason1981compliance}. A short window of wrist samples before each descent gives a baseline wrench $\mathbf{w}_0$ that cancels the grasped object's weight. The arm then descends along world $-z$, at fixed rotation $\boldsymbol{\theta}$ using the Pilz LIN Cartesian-linear planner~\cite{pilz2020planner}, and aborts once the world-frame vertical force delta, $\Delta F_z^{\text{world}}$ exceeds a threshold $F_z^\text{thresh}$. The contact height is then logged, and the gripper releases the object.

\subsubsection{Post-Release Consolidation Push}

After releasing the object, two independent axis-aligned pushes ($-x$ and $-y$) consolidate it toward the workspace origin, absorbing the small lateral clearance left by the contact-safe descent together with the residual drift between the planned and achieved poses. Each axis is skipped if the object is already at the origin wall, if the gripper start pose would lie outside the container, or if the corridor behind the object is already occupied. On a surviving axis, the gripper descends low on the object's side at a low height chosen to push it translationally rather than tip it, then executes a slow Cartesian-linear push that aborts once $\|\Delta\mathbf{F}_{xy}^{\text{world}}\|$ exceeds a small threshold. This gives a deterministic stop on a wall or a neighboring object.

\section{Experiments}
\label{sec:experiments}

We evaluate the full pipeline on a real Franka arm on a 3D-printed object set and a YCB subset. We also conduct ablation study isolating gripper-aware optimization, the consolidation push, and mesh-derived geometry. And finally we compare our method against a baseline of prior irregular-object packing work. 

\begin{figure}[h]
\centering
\settoheight{\imgheight}{\includegraphics[width=0.68\textwidth]{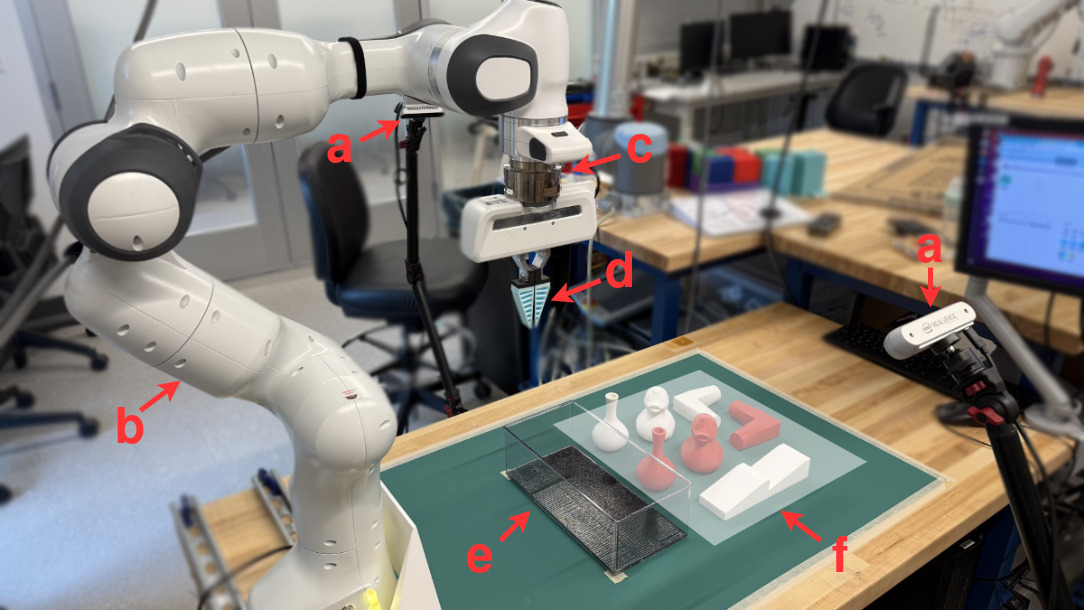}}%
\includegraphics[width=0.68\textwidth]{media/isrr_experiment_setup.jpg}%
\hspace{0.02\textwidth}%
\includegraphics[height=\imgheight]{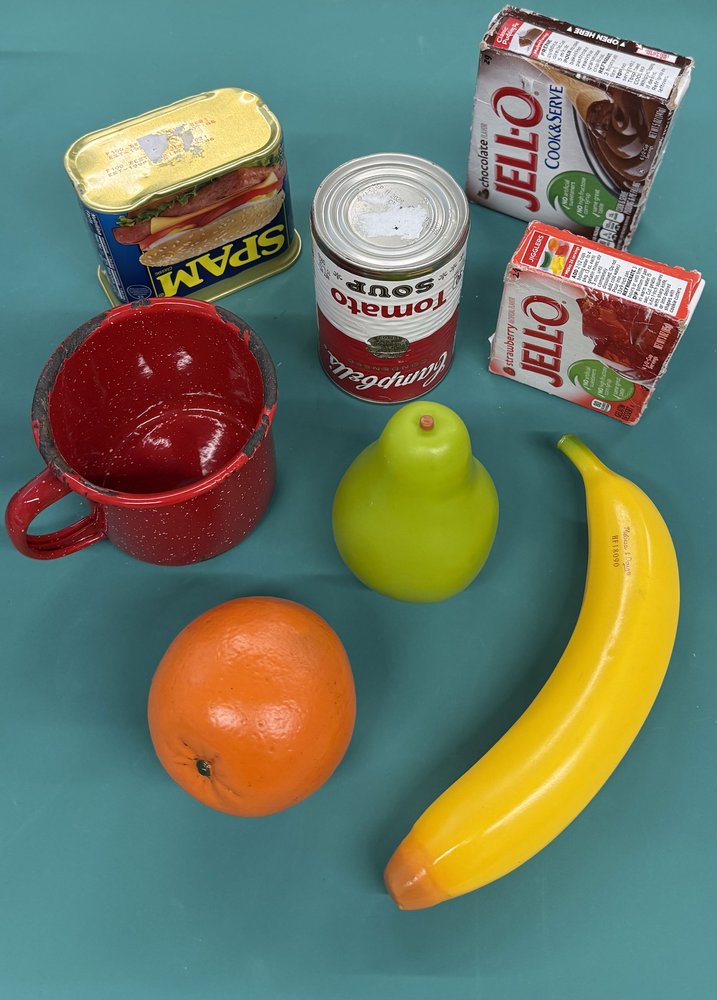}
\caption{Hardware platform and object set. $\mathbf{Left}$: (a) Intel RealSense D415 and D455 depth cameras mounted on opposite sides of the workspace, (b) 7-DoF Franka Emika Panda~\cite{haddadin2022franka}, (c) ATI wrist force/torque sensor, (d) 3D-printed soft parallel fin-ray gripper, (e) Container, (f) Staging area with the seven-object set. $\mathbf{Right}$: the eight-object YCB subset used to test generalization.}
\label{fig:object_set}
\end{figure}

\subsection{Hardware and Object Sets}
\label{sec:objects}

The platform is shown in Fig.~\ref{fig:object_set}. The ATI wrist force/torque sensor streams at 100\,Hz, and the container has inner dimensions $6 \times 12 \times 4$\,in. Optimization and motion planning run on an NVIDIA RTX 3090 workstation using MoveIt~\cite{coleman2014moveit} with the Pilz industrial motion planner~\cite{pilz2020planner}. We use a set of seven 3D-printed replicas of irregular household and warehouse objects selected to span the variation that drove our formulation. The set contains one flat support that anchors the base layer, two tall and slender vases with a tight approach angle tolerance, two duck-shaped objects with curved geometry, and two L-shaped objects whose concave profile rewards careful orientation. Each entry comes with a CAD mesh, a sampled surface point cloud, and a single pre-registered antipodal grasp in the object's own frame (Fig.~\ref{fig:object_set}).

To test generalization beyond this set, we additionally evaluated our approach on an eight-object subset of the YCB Object and Model Set~\cite{calli2017ycb}: banana, gelatin box, mug, orange, pear, potted meat can, pudding box, and tomato soup can (Fig.~\ref{fig:object_set}) . These objects 
span a comparable range of shape complexity, from simple prisms to curved and asymmetric geometry. 

\subsection{Perception}
\label{sec:perception}

Two RealSense cameras are mounted on opposite sides of the workspace reduce occlusion. For each camera, depth measurements are averaged over ten captures. Camera poses relative to the robot base are estimated from a fixed workspace ArUco marker~\cite{garridojurado2014aruco} using SQPnP~\cite{terzakis2020sqpnp}, and repeated estimates are averaged in $SE(3)$. The table is removed by a RANSAC plane fit~\cite{fischler1981ransac}, the remaining points are clustered with DBSCAN~\cite{ester1996dbscan} to isolate the target object, and the cluster is matched to the CAD library by point-to-point ICP~\cite{besl1992icp} scored by Chamfer distance. Combining object at the matched pose with the gripper at the pre-registered grasp gives the composite body directly in the robot base frame.

\subsection{Test Sequences}
\label{sec:sequences}

Objects are sequenced by decreasing spatial extent,
\begin{equation}
s_k \;=\; \max(\text{dim}(P_k)) \cdot V_\text{AABB}(P_k),
\label{eq:sequencing}
\end{equation}
which biases picking large flat objects so that the base layer is formed first. 

To isolate the contribution of the sequencing rule, we evaluate the full system on three fixed orderings of the object set, each repeated for five trials. The \emph{Optimized} sequence follows the largest-first ordering rule of Eq.~\eqref{eq:sequencing}. The \emph{Random 1} and \emph{Random 2} sequences are random orderings that expose how much density the optimizer can achieve without the benefit of the sequencing rule. We additionally run the YCB object set on a fixed sequence also for five trials, without random orderings, since the goal is to test generalization rather than to isolate the sequencing rule again. The sequences are listed in Table~\ref{tab:sequences}.


\begin{table}[t]
\caption{Test sequences (five trials each).}
\label{tab:sequences}
\centering
\begin{tabular}{p{0.22\columnwidth} p{0.68\columnwidth}}
\toprule
Sequence & Object ordering \\
\midrule
Optimized & support, duck, duck, vase, vase, L, L \\
Random 1 & vase, L, vase, L, duck, duck, support \\
Random 2 & duck, vase, L, support, L, vase, duck \\
YCB & mug, tomato soup can, banana, potted meat can, pudding box, gelatin box, pear, orange \\
\bottomrule
\end{tabular}
\end{table}

\subsection{Evaluation Metrics}
\label{sec:metrics}

We report metrics in three groups. Packing-quality metrics describe the final static packing result in the container, execution-robustness metrics describe the run-time behavior of the pipeline, and computational metric describes the cost of producing each placement.

\subsubsection{Packing Quality.}

\emph{Space utilization.} Ratio of total volume of placed objects to the total volume of the object set. The container is sized so the full set only barely fits, which keeps space utilization a discriminative metric. An object is counted as placed only if it lies fully within he container planform up to the vase height. An object can only be counted in a binary in/out sense, since partial overhangs are not realistic for downstream warehouse handling. This is a variant of the density metric used by prior bin-packing work~\cite{zhao2022pct,pan2023sdfpack,wang2022densepacking}.

\emph{Number of objects packed.} The number of objects successfully placed inside the container by the end of a run, or before an abort. Reported with space utilization to distinguish dropping one large object vs. several small ones.

\subsubsection{Execution Robustness.}

\emph{End-to-end success rate.} The fraction of orderings in which every object is placed without an aborted insertion, dropped grasp, or toppled neighbor. A single failure invalidates the ordering. This is the closest analogue to the order-level success rate reported by Wang and Hauser~\cite{wang2022densepacking}.

\emph{Per-placement success rate.} For each attempted placement we record either success or failure due to optimizer-infeasible, grasp failure, approach collision, insertion abort, or post-release disturbance.

\subsubsection{Computational Cost.}

\emph{Optimizer wall time.} Mean and maximum time spent inside the CMA-ES loop per object, reported separately from end-to-end iteration time because the latter is dominated by robot motion.

\subsection{Ablations}
\label{sec:ablations}

We isolate the contribution of each of the three pipeline functions (i.e., using the 3-D irregular shape of each object directly, pushing, and gripper-aware optimization) through ablation study against the full system. Each ablation toggles a single function and leaves everything else unchanged. All ablations run on the \emph{Optimized} sequence defined in Sec.~\ref{sec:sequences}. Variants are summarized in Table~\ref{tab:ablations}.

\begin{table}[h]
\caption{Ablation variants, run on the \emph{Optimized} sequence.}
\label{tab:ablations}
\centering
\begin{tabular}{p{0.08\columnwidth} p{0.32\columnwidth} p{0.55\columnwidth}}
\toprule
ID & Ablations & Hypothesis tested \\
\midrule
A0 & Full system & Reference for paired comparison. \\
A1 & Prism-like geometry & Planform-extrusion proxy~\cite{cao2026tase} cannot pack as dense as using the irregular shapes directly. \\
A2 & No consolidation push & Post-release push contributes measurably to final density.\\
A3 & No gripper-aware optimization & Omitting the gripper from the composite body causes collision. \\
\bottomrule
\end{tabular}
\end{table}

A1 swaps each object's mesh-derived sphere tree for a prism extruded from its XY planform. A2 disables the consolidation push. A3 removes the gripper points $P_G$ from the composite body and drops the orientation constraint, though the gripper stays attached during execution so approach-time interference is still exposed. Each variant is run for five trials.

\subsection{Comparative Study}
\label{sec:comparative_setup}

To evaluate our approach relative to prior work, we additionally compare against Wang and Hauser's heightmap-minimization (HM) formulation~\cite{wang2022densepacking}, the closest existing method targeting dense packing of irregular objects with a real manipulator. We hold perception, grasp selection, and execution identical to our own pipeline and substitute our optimizer with Wang and Hauser's own grid search over discretized $(x,y)$ positions and angles, scored by minimizing the resulting heightmap. Since its quality depends on grid resolution, we evaluate two settings, five trials each: \emph{HM\_Low\_Resolution} reproduces their reported resolution (Table~\ref{tab:hm_compare}), and \emph{HM\_High\_Resolution} matches the candidate count of our own CMA-ES budget, isolating whether any shortfall traces to a coarse search or to the grid-based formulation itself.

\section{Results and Discussion}
\label{sec:results}

We organize the results around three questions. Sec.~\ref{sec:res_real} reports the full system across the four test sequences and quantifies the effect of sequencing. Sec.~\ref{sec:res_ablation} isolates the contribution of mesh-derived geometry, the post-release push, and gripper-aware optimization through ablations on the \emph{Optimized} sequence. Sec.~\ref{sec:res_comparative} shows how our method performs compared to the baseline. A video recording of the test sequences, each ablation case, and the compared baseline, is provided as supplementary material.

\subsection{Packing Across Test Sequences}
\label{sec:res_real}

\subsubsection{3D-printed Objects}
\begin{figure*}[t]
\centering
\includegraphics[width=\textwidth]{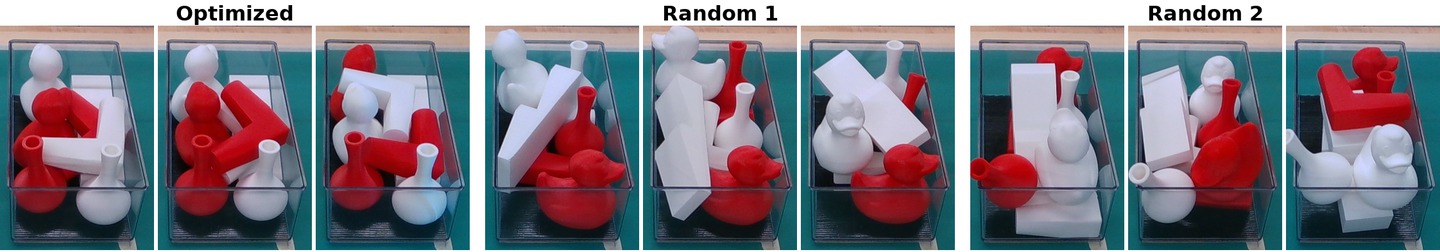}
\caption{Three final packing results (out of five) for each sequence. The Optimized sequence produces visibly flatter and tighter configurations, while the random orderings leave larger residual cavities as small objects are committed early.}
\label{fig:test_final}
\end{figure*}

Table~\ref{tab:real} reports the full system on the three test sequences of Sec.~\ref{sec:sequences}, each repeated for five trials. The key result is that 14 out of 15 orderings completed end-to-end. The single failure occurred in the Random 1 sequence, where the flat support object was not picked up due to perception drift and the run terminated after six objects. Final packing results for each sequence are shown in Fig.~\ref{fig:test_final}, and a per-configuration summary of utilization and objects packed is shown in Fig.~\ref{fig:config_bars}(a). 

\begin{table}[t]
\caption{Average system performance of three test sequences (five trials each).}
\label{tab:real}
\centering
\small
\begin{tabular*}{\textwidth}{l@{\extracolsep{\fill}}c@{\extracolsep{\fill}}c@{\extracolsep{\fill}}c@{\extracolsep{\fill}}c}
\toprule
Metric & Optimized & Random 1 & Random 2 & Overall \\
\midrule
Space utilization (\%)        & $\mathbf{95.59 \boldsymbol\pm 6.06}$ & $81.37 \pm 0.02$ & $75.19 \pm 5.66$ & $84.05 \pm 9.89$  \\
Objects packed                & $\mathbf{6.60 \boldsymbol\pm 0.55}$ & $6.00 \pm 0.00$ & $5.40 \pm 0.55$ & $6.00 \pm 0.65$ \\
End-to-end success            & $\mathbf{5/5}$             & $4/5$             & $\mathbf{5/5}$             & $14/15$ \\
Per-placement success         & $\mathbf{35/35}$           & $34/35$           & $\mathbf{35/35}$           & $104/105$ \\
CMA wall time (s/obj)         & $3.56 \pm 0.20$ & $3.32 \pm 0.21$ & $\mathbf{3.19 \boldsymbol\pm 0.11}$ & $3.34 \pm 0.22$ \\
\bottomrule
\end{tabular*}
\end{table}

\subsubsection{YCB Objects}
On the YCB sequence (Table~\ref{tab:ycb}), the system completes all five trials end-to-end with every attempted placement succeeding ($40/40$).

\vspace{-2em}
\begin{table}[h]
\caption{Average system performance on the YCB objects (five trials each).}
\label{tab:ycb}
\centering
\small
\begin{tabular*}{\textwidth}{l@{\extracolsep{\fill}}c@{\extracolsep{\fill}}c@{\extracolsep{\fill}}c}
\toprule
 & Space utilization (\%) & Objects packed & CMA wall time (s/obj) \\
\midrule
YCB subset & $82.45 \pm 10.36$ & $6.60 \pm 0.89$ & $3.00 \pm 0.46$ \\
\bottomrule
\end{tabular*}
\end{table}

\vspace{-3em}
\begin{figure}[H]
\centering
\settoheight{\imgheight}{\includegraphics[width=\textwidth]{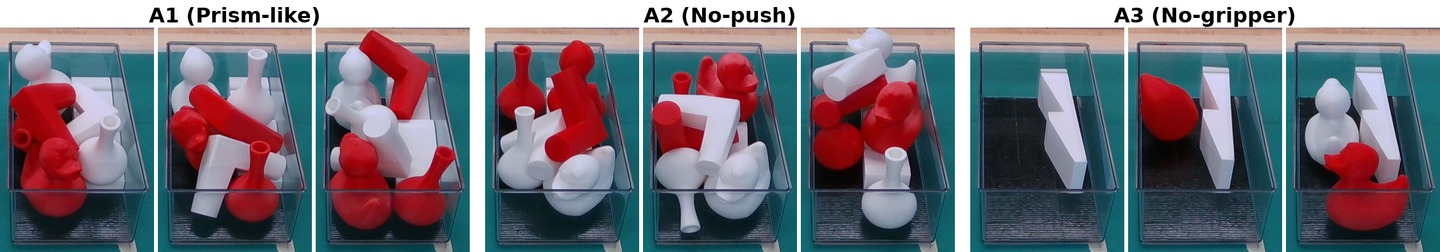}}%
\includegraphics[height=1.15\imgheight]{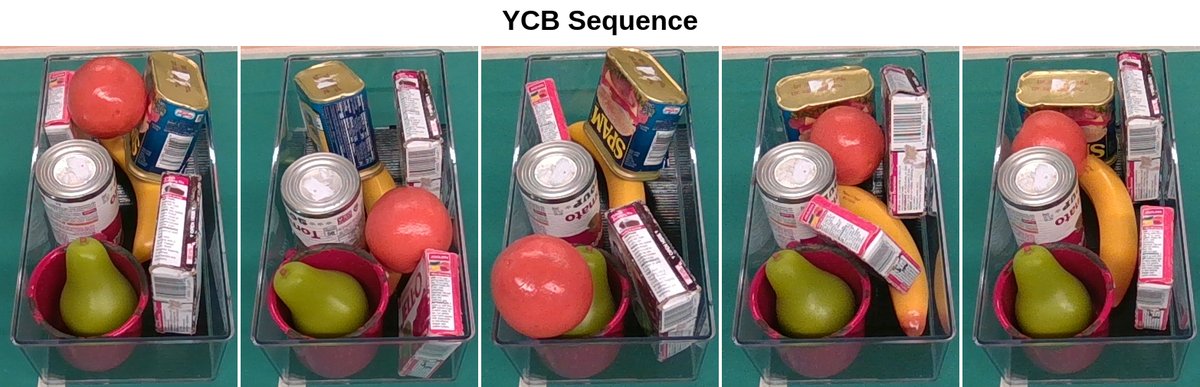}
\caption{Final packing results for all five trials of the YCB sequence.}
\label{fig:ycb_sequence}
\end{figure}

\paragraph{Sequencing sets an upper bound on achievable density.}
The roughly 20-point drop between Optimized and Random~2 exceeds the within-sequence variation by a wide margin, so sequencing has a real effect on achievable density. Even when small or concave objects are committed early, the optimizer still returns reachable, collision-free placements for every subsequent object, but cannot recover the volume lost to a base layer that was never built. End-to-end success and CMA wall time are essentially unaffected, so this cost is purely in density, not reliability or search effort. The YCB result lands in the same Random~1/Random~2 range of the 3D-printed set. The fact that it reaches comparable density and preserves perfect per-placement success suggests the robustness of our composite-body formulation and packing terms on a different object set.

\subsection{Ablations}
\label{sec:res_ablation}

\begin{figure*}[h]
\centering
\includegraphics[width=\textwidth]{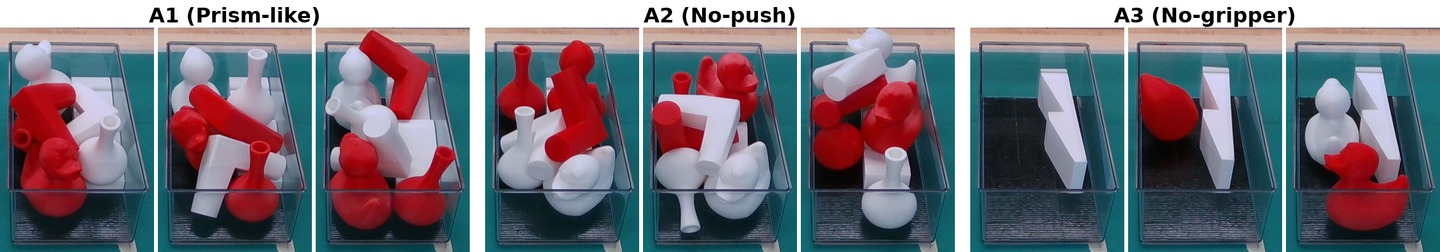}
\caption{Three final packing results (out of five) for each ablation variant on the Optimized sequence.}
\label{fig:ablation_final}
\end{figure*}

\begin{figure}[h]
\centering
\includegraphics[width=\columnwidth]{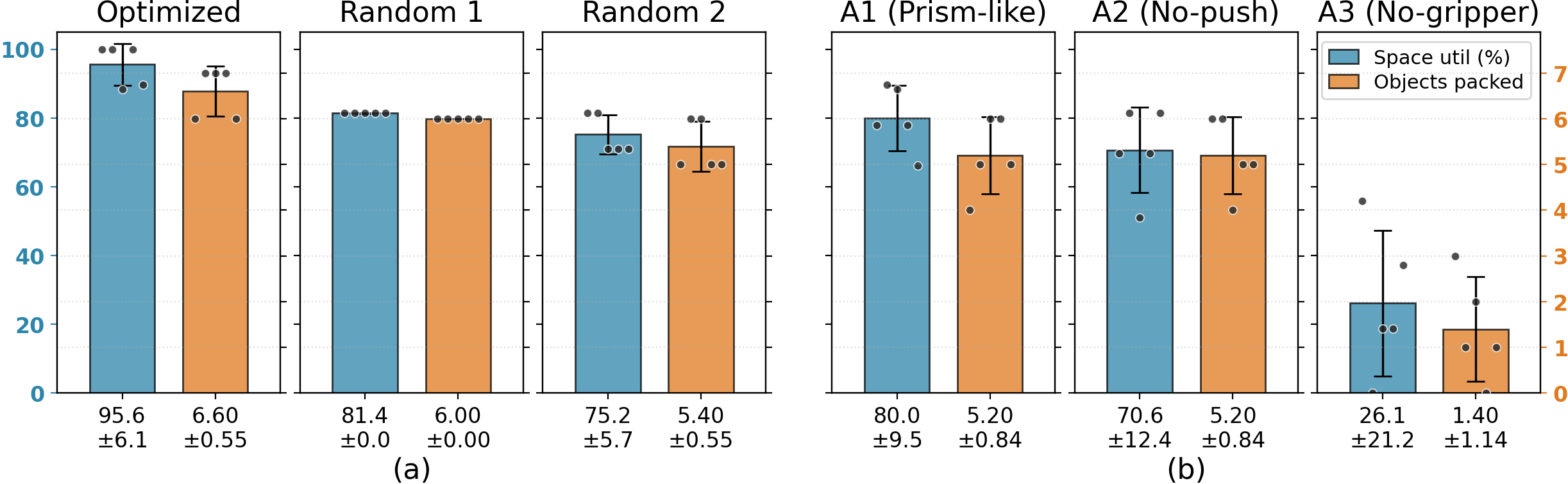}
\caption{
Comparison of space utilization and objects packed. (a) The left three bars correspond to the sequential packing tests. (b) The right three bars correspond to the ablation variants on the \emph{Optimized} sequence. Bars show the mean over five trials per configuration with error bars indicating one standard deviation.}
\label{fig:config_bars}
\end{figure}

We compare the full system against three ablations on the \emph{Optimized} sequence, with five trials each. Table~\ref{tab:ablation_results} reports the result. Fig.~\ref{fig:ablation_final} shows a representative final packing result per variant, and Fig.~\ref{fig:config_bars}(b) shows the ablation bars next to the sequencing bars of (a) on the same scale.

\vspace{-2em}
\begin{table}[h]
\caption{Paired ablation on the \emph{Optimized} sequence.}
\label{tab:ablation_results}
\centering
\small
\begin{tabular}{lcccc}
\toprule
Metric & A0 (Full) & A1 (Prism-like) & A2 (No-push) & A3 (No-gripper) \\
\midrule
Space utilization (\%)        & $\mathbf{95.59 \boldsymbol\pm 6.06}$ & $80.00 \pm 9.50$ & $70.62 \pm 12.43$ & $26.09 \pm 21.23$ \\
Objects packed                & $\mathbf{6.60 \boldsymbol\pm 0.55}$ & $5.20 \pm 0.84$  & $5.20 \pm 0.84$  & $1.40 \pm 1.14$ \\
End-to-end success            & $\mathbf{5/5}$  & $\mathbf{5/5}$   & $\mathbf{5/5}$   & $0/5$ \\
Per-placement success         & $\mathbf{35/35}$ & $\mathbf{35/35}$ & $\mathbf{35/35}$ & $7/35$ \\
CMA wall time (s/obj)         & $3.56 \pm 0.20$ & $\mathbf{1.87 \boldsymbol\pm 0.12}$  & $3.30 \pm 0.28$  & $2.69 \pm 0.39$ \\
\bottomrule
\end{tabular}
\end{table}

\vspace{-2em}

\emph{No gripper-aware optimization (A3)} collapses the pipeline. Space utilization drops significantly, only $1.4$ objects are placed on average, and no trial completes end-to-end ($0/5$). Two failure modes account for the collapse. First, with the gripper removed from the composite body, the optimizer clips the first support object flush against the container origin, so during execution the fingers strike the two adjacent walls and the object rolls free into the container before release. Second, because the orientation term acts on the gripper axis, removing the gripper from the optimization also removes the upright preference, and the optimizer returns physically impossible upside-down poses for asymmetric objects such as the duck. CMA wall time drops to $2.69 \pm 0.39$\,s/obj because the search terminates as soon as the object-only sphere tree fits, but the resulting poses are unreachable for the real arm-and-gripper system.

\emph{No consolidation push (A2)} causes the larger density drop among the variants that still complete end-to-end, from 95.6\% to 70.6\% utilization at unchanged 35/35 per-placement success.

\emph{Prism-like geometry (A1)} reaches $80.0\%$ utilization and $5.2$ objects packed, both below the full system, while CMA wall time roughly halves. 

\vspace{-2em}

\paragraph{The gripper belongs in the planner, and pushing recovers what it costs.}

The A3 result reflects a qualitative failure rather than a gradual loss of density. The gripper lets the optimizer verify that the arm and gripper together can physically reach a placement, not merely that the placement suits the object alone. Removing it from the composite-body representation eliminates that check entirely, so the optimizer can no longer tell whether its output is executable. The drop observed under A2 reflects the price of that reachability. Since gripper-aware optimization must leave lateral clearance around a neighboring object to keep the gripper collision-free during approach, this clearance persists even after the gripper retracts. The consolidation push then closes that clearance after release, which makes the division of labor explicit. Planning secures reachability at the cost of clearance, and pushing subsequently recovers the density that this clearance leaves behind.

\paragraph{Geometric fidelity vs.\ search cost.}
A1's lower wall time follows from its smaller, shallower sphere tree, and its utilization appears competitive because planform extrusion preserves each object's footprint. Curved and concave objects, however, are seated against an idealized prism rather than against their true shape. As a result, the optimizer reserves more space than the object actually occupies, and neighboring objects cannot nest into its concavities. This effect is most visible for the L-shaped object, whose true volume is far smaller than that of its prism hull. Planform extrusion therefore remains workable for catalogs of upright, prism-like items~\cite{cao2026tase}, but it breaks down for irregular or concave ones.

\subsection{Comparative Study}
\label{sec:res_comparative}

\begin{figure*}[h]
\centering
\includegraphics[width=\textwidth]{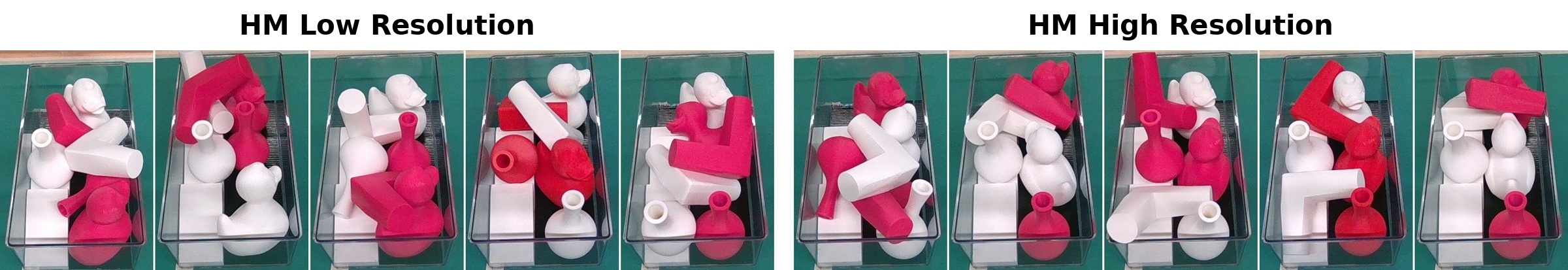}
\caption{Final packing results for the HM baseline on the Optimized sequence.}
\label{fig:hm_resolution_comparison}
\end{figure*}

Table~\ref{tab:hm_compare} and Fig.~\ref{fig:hm_resolution_comparison} report the comparison on the \emph{Optimized} sequence. Our system achieves the highest packing quality of the three configurations, exceeding both HM baselines in space utilization and in the number of objects packed. All three configurations complete every trial end-to-end ($5/5$) with a clean per-placement record ($35/35$), so the gap lies entirely in how densely each method packs the container rather than in failed grasps or collisions.

\vspace{-2em}
\begin{table}[h]
\caption{Full system (Optimized sequence) vs. Wang and Hauser~\cite{wang2022densepacking} heightmap-minimization (HM) baseline, 
with two search budgets: HM\_Low\_Resolution (10\,mm XY step, $\Delta r = \pi/4$, their Sec.\ VI setting) and HM\_High\_Resolution (5\,mm step, 16 yaws, a matched search budget). Five trials each.}
\label{tab:hm_compare}
\centering
\small
\begin{tabular*}{\textwidth}{l@{\extracolsep{\fill}}c@{\extracolsep{\fill}}c@{\extracolsep{\fill}}c}
\toprule
Metric & Optimized & HM\_Low\_Resolution & HM\_High\_Resolution \\
\midrule
Space utilization (\%)        & $\mathbf{95.59 \boldsymbol\pm 6.06}$ & $77.66 \pm 8.33$ & $82.35 \pm 6.06$ \\
Objects packed                & $\mathbf{6.60 \boldsymbol\pm 0.55}$  & $5.00 \pm 0.71$  & $5.40 \pm 0.55$ \\
\bottomrule
\end{tabular*}
\end{table}

\vspace{-2em}

\vspace{-2em}
\subsubsection{Algorithmic and Computation Comparison}

Our optimizer searches its 5-DOF pose continuously (Eq.~\eqref{eq:searchvec}), whereas HM enumerates a discrete grid of positions and angles and scores each heightmap cell.
HM's quality is a function of grid resolution. The gain from HM\_Low\_Resolution to HM\_High\_Resolution comes entirely from using a finer grid, whereas our continuous search has no such parameter to tune and still reaches higher utilization than either discretization. 

There is also a scoring bias from rasterization. HM scores a yaw by rasterizing the object's rotated footprint onto the grid, so for a radially asymmetric object the score depends on how the footprint boundary aligns with the grid lines rather than on the true occupied area. This biases the search toward certain orientations in a way that sampling more yaws cannot remove. Our optimizer instead scores orientation directly on the sphere-tree geometry (Sec.~\ref{sec:objective}) without rasterizing, so it is not subject to this bias. 

In addition, refining HM's grid increases the number of candidates polynomially. Wall time rises from $0.42\pm0.03$ to $1.04\pm0.03$\,s/obj for that gain in utilization, and closing the remaining gap to our own result would require much larger refinement at correspondingly larger cost. 

\subsection{Limitations}
\label{sec:limitations}

Our results assume a pre-registered library with a single antipodal grasp per object. This matches industrial unit-catalog settings but excludes novel-object packing, where grasp planning is part of the inner loop. The system also does not re-estimate each object's pose after placement, instead treating the raw merged point cloud inside the container as $\mathcal{C}_k$. Sensor noise and partial occlusion therefore inflate $\mathcal{C}_k$ relative to the true placed geometry, which slightly restricts the optimizer's view of free space. 


\section{Conclusions}
\label{sec:conclusion}
We presented a closed-loop, real-time pipeline for dense packing of irregular objects with a real manipulator, treating the object and gripper as a single composite body optimized over five degrees of freedom on a GPU and absorbing perception and contact drift through force-monitored descent, a post-release consolidation push, and re-perception between placements. We tested our method 
using a real robot to pack a set of flat, curved, and concave objects, including generalization to a YCB subset. The system completed 19 out of 20 trials end-to-end and achieved 95.6\% space utilization on the Optimized sequence, exceeding a baseline at both tested grid resolutions and at a fixed computational cost. Ablations show that gripper-aware optimization governs whether a placement is reachable rather than how dense it is, the consolidation push recovers the clearance that gripper-aware planning leaves behind, and non-prism object representation is necessary to achieve high packing density for non-compact shapes. Extensions include re-estimating placed object  poses to refine the container status and lifting the pre-grasp assumption toward novel-object packing.

\section*{Acknowledgment}

This work was funded by an Amazon Robotics$-$WPI grant and supported in part by the National Science Foundation under the award ECCS-2338703. 

\newcommand{\bibvenue}[2]{%
  \ifcsname bibvenue@#1\endcsname
    #1%
  \else
    #2 (#1)%
    \expandafter\gdef\csname bibvenue@#1\endcsname{}%
  \fi
}

\bibliographystyle{Springer_style/bibtex/splncs03}
\bibliography{ref}

\end{document}